# A survey of SMS based Information Systems

Manish R. Joshi[1] Varsha M. Pathak[2]

[1]School of Computer Science, North Maharashtra University, Jalgaon, MS, India
joshmanish@gmail.com

[2]MCA department IMR Jalgaon, North Maharashtra University Jalgaon, MS, India,
pathak.vmpathak.varsha@gmail.com,

***Abstract—*** Short Message Service (SMS) based Information Systems (SMSbIS) provide an excellent alternative to a traditional approach of obtaining specific information by direct (through phone) or indirect (IVRS, Web, Email) probing. Information and communication technology and far reaching mobile penetration has opened this new research trend Number of key players in Search industry including Microsoft and Google are attracted by the expected increase in volume of use of such applications. The wide range of applications and their public acceptance has motivated researchers to work in this research domain. Several applications such as SMS based information access using database management services, SMS based information retrieval through internet (search engine), SMS based information extraction, question answering, image retrieval etc. have been emerged.

With the aim to understand the functionality involved in these systems, an extensive review of a few of these SMSbISs has been planned and executed by us. These systems are classified into four categories based on the objectives and domains of the applications. As a result of this study a well structured functional model is presented here. The model is evaluated in different dimensions, which is presented in this paper. In addition to this a chronological progress with respect to research and development in this upcoming field is compiled in this paper.

Such an extensive review presented in this paper would definitely help the researchers and developers to understand the technical aspects of this field. The functional framework presented here would be useful to the system designers to design and develop an SMS based Information System of any specific domain.

***Keywords-component*** -SMS based Information System, Information Retrieval, Functional Model, Function Point, M-Goveranance, Research Prototype, Research Models.

## 1. INTRODUCTION

Short Message Service (SMS) is understood as one of the revolutionary development of the Information and Communication Technology (ICT). The increasing data transfer rates of mobile technology has complimented this development with a great extents. People irrespective of their socio-economic background use mobiles as information access terminals. It has become the most reachable bridge to link up the digital divide. The underline power of data exchange using SMS service is playing a key role to evolve the applicability of mobiles in information systems and retrieval. With this background a comprehensive research and development have been already started by the researchers. Number of SMS based Information Systems (SMSbIS) are being developed by the researchers, developers and professional vendors.

In any generic SMSbIS, a text query in the form of SMS is send from mobile to an appropriate server of the service provider. The query is received by the server then processed and relevant data, text or image is accessed from the available data source like database or the WEB . The resultant data/text and even image is then returned back as a response to the sender’s mobile unit.

The enhanced development of mobile technology into 3G technology has come up with fast internet access , multimedia data transfers, video SMS and high data transfers rates. The penetration of mobile networks upto remote places and natural language support has made “SMS based information retrieval” as a new challege to the researchers and developers.

With the prime interest to initiate self contribution in the research related to SMS based information systems, the authors have investigated the current developments in the field. They found a profound work has been already started. The developments in applied and core research associated in this field are thus needed to be reviewed. Understanding this need the authors of this paper have selected a few systems from various domains. Based on the related published work an extensive review has been planned. The objectives set in this study are listed below.

- Classify the systems based on the objectives and domains of the applications.
- Identify various Functional features of the systems.
- Cross verify the systems of different classes and reveal the differences in their functional aspects.
- Present chronological comparison of the systems to understand the development in the field from 2001.
- Present open issues of SMSbIS Research and Development.

The sections written in this paper are as briefed here. The systems are described separately in annexure in tabular form. The classification of systems and functional model are explained in the section II. Details of the various Functional Features in concern with the five Functional Categories are explicitly described from section III to section VI. The Functional Features are found to be supported with various technical options by different systems. These related technical options are termed as function points. These Function Points as experimented by the systems are presented in tabular form in respective sections. These tables are analysed in respective sections. Other perspective of the evaluation is to look at the chronological progress in the reasearch of this field. For this the systems are sorted on base of year of their publications from year 2001 to 2013. This chronological analysis in presented in Section VIII. At last conclusions are submitted covering research related open issues in section IX.

## 2. Systems Models.

When we evaluated the systems to understand the functional aspects, we found they represent either the applied research work or the core research. Both types of systems are developed by the developers, researchers of industries, academics and government sectors. Thus it is necessary to distinguish these systems on the basis of their domains with respect to the goals set for their development. Analysis of the functional model with respect to these identified system classes is then used to extract the open issues in this research fields. The classification of these systems and the functional model is presented in this section as follows.

### *A. System Classification*

Due to increasing applicability of SMSbISs, different systems across various domains are available. From the available literature we selected Forty three such systems for the systematic study.
These selected systems are classified in four different categories as below, depending on the identified domains of the applications.

- Professional Systems: Few software/hardware vendors are coming up with products to support the SMS based information access. Problems related to e-Governance, business process and banking sectors are experimented in these systems. Study of these systems reveals the development of functionality in professional's point of view. Six practically implemented solutions developed by the professional vendors are selected to represent this class. All of these systems belong to applied work. This class is named as 'Category A' for further reference.

- M-governance Systems: Both the applied and core research is in progress for the effective development of m-Governance projects. The wide acceptance of mobiles as information access terminals has initiated the research in various M-government proposed projects. Six such proposed works are selected for the study. These models are proposed to develop solutions with Government to Citizen (GTOC) information exchange using mobiles. Both the professionals and the researchers contribute in the development of such M-governance proposals. This class is named as 'Category B' for further reference.

- Research Prototypes: From a wide range of systems with core research aspects. Around 16 such systems are selected under this class. Various technical options related to SMS based query processing and information access could be found in these systems. Innovative ideas and methodologies can be found in these systems. These systems are developed to handle issues related with Natural Language Query (NLQ). Thus we found these systems as very interesting to study the NLP related Functional features. We call this group as 'Category C'.

- Research Models: Research models for SMS based Information systems, are being widely developed by various researchers from universities all over the world. These systems are considered under applied research. 15 such research models are referred in this paper. The related published work has revealed wide range of technical aspects of these types of applications. Major domain that is being governed by most of these systems is the education domain. We refer this class as 'Category D'.

### *B. Functional model*

To investigate the underline technology various functional components of the selected SMS based information systems have been studied in detail. A functional model is framed out of this study. Five functional levels are assigned from the overall functionality involved in these systems viz. *Query Construction Interface, Query Processing, Core Service Logic, Security & Privacy and Response Generation.* These levels of functionality are here termed as "Functional Categories". These Functional Categories are meeting with the corresponding level of functionality with the set of sub-functions. These sub-functions will be referred as "Functional Features" in this functional model.

This initially set functional model is depicted in Fig. 1. It is further explored by evaluating each system for this set of functional features. In this evaluation, practical implementation of corresponding sub-functions is understood. These functional options are recognized as "Function Points", in this framework.
This functional model is thus consisting of the five "Functional categories" and collectively twenty three "Functional features" from these functional categories. In this survey the estimated "Function Points" used to implement each functional feature, are composed into corresponding sections with reference of the supporting systems. This information is then summarized in tabular form in respective sections. Each table of respective section corresponds to the specified functional category. Each table consists of three major columns. First column corresponds to the Functional features. Second column list outs “Function Points”. The third Column gives collective functional view of the forty three systems divided into four categories. This view consists of a matrix (Function Points *x* Systems). It assigns a check mark (√) or empty cell (Space) as the matrix elements. A check mark means that a system S# supports the functional issue with the function points M#. The blank cell in the table indicates that the system S# does not support the option M#. If all the cells in the system S#’s column for a particular functional issue are blanks, it means the system does not support the functionality at all. It also means that the published work of the system has not given any reference of that functional issue in its document. The authors have associated the function option with the system only if the corresponding research has mentioned it clearly in its published work.
It became easier to analyze these tables independently to extract the open issues related with the concerned set of functionality.

## 3. QUERY CONSTRUCTION INTERFACE (QCI)

User needs a suitable interface to construct SMS query. Twelve Functional points are distributed among Five functional features in this category. Editing SMS, voice and video messages, natural language barrier and query formats are the related concerns elaborated one by one in the first part of this section. Table I, in first subsection summarizes different function points with reference to the Forty three selected systems. This table is then analyzed to present the observations in next subsection.

### 3.1 QCI functional features

With reference of various systems five functional features are identified in this functional category. Each feature is explored independently within individual systems to extract the method used by the system to support the feature. Table – I gives this cross view of thirteen different Function Points, across these five features vs. the forty three systems. Let us first look at the features and the respective Functions.

*A. SMS edit*

Editing SMS on mobiles is the beginning of the whole communication aspect. 4 different options of implementing this feature are found from the study. It is found that most of the mobiles’ software provides a usable **Text editor**. The most important precondition of query interface is the small space of the screens of mobiles / PDA units; this formally limits the number of characters to be used (Near about 160 characters). A suitable query formation interface with this restriction is the important feature of the systems. Pratima Mitra et. Al. of System S13 has well discussed this fact. With reference to Abdul-Kareem [12], System S9 enlists various concerns of query interface. System S9 supports special query formation interface. System S5 uses billing client **template** to enter query of meter reading. Some predefined services like that of railway information system may be supported with **menu** to construct the query. Users need to select through the set of the terms in menu to compose the query. Modern mobiles support **Browsers** [24], [34]. Systems S19 and S29 are the respective references those use these web browsers to send query request to the server. **Voice SMS-**People prefer to talk rather than type. It is the fact that voice conveys more information than text. Voice message of 30 to 60 seconds carry more than 160 character text information (Jeff Epstien, “Voice SMS” an article in interview). Voice recognition mechanism in mobiles will be useful to convert voice commands to text messages. Interactive voice recorder service (IVRS) is one of the special features of Value added services of modern wired and wireless telecommunication technology. Auto-Assist is the advanced application of Voice SMS.

*B. Image/MMS*

3G technology has encouraged Image based messages. Practically if MMS are included in the query to send multimedia information then more wide implementations are possible. The paper [6] has enlisted few applications such as “Fire fighters” where Video SMS are useful. Mobile based “Window Shopping” is experimented in system S15. System S24 uses MMS for sending and receiving voter’s photo from polling stations. System S26 suggests an interesting application of tourism domain. The system lets the tourists to send images as MMS, of any monuments they watch when they visit such place and can receive information as SMS related to that monument. S36 gives recap of how images of ECG, Heart images, Lungs Images could be send as MMS query to receive the diagnostic information in Mobile based Health Care systems. S41 shows image based information retrieval using Android mobiles. The mobile is used as client interface for sending Image query to access information from the server by applying image processing algorithms.

*C. Language Support*

Better applicability of SMS based Information systems (SMSbIS) can be achieved by resolving language barrier. ‘Language support’ is thus an unavoidable feature. Users can meet flexibility in query construction if natural language support is available. SMS NLQ processing has become an emerging research area now.
*NLP approaches are under progress in this respect [15, 16]. In system S8 for example, ontology based natural query language is experimented. NLQ based systems such as (S8, S12) have focused on format free NLQs using single language (English). Apart from the semantic resolutions, the SMS formation has added problem of new abbreviations (for : 4, letter : l8r etc.).*

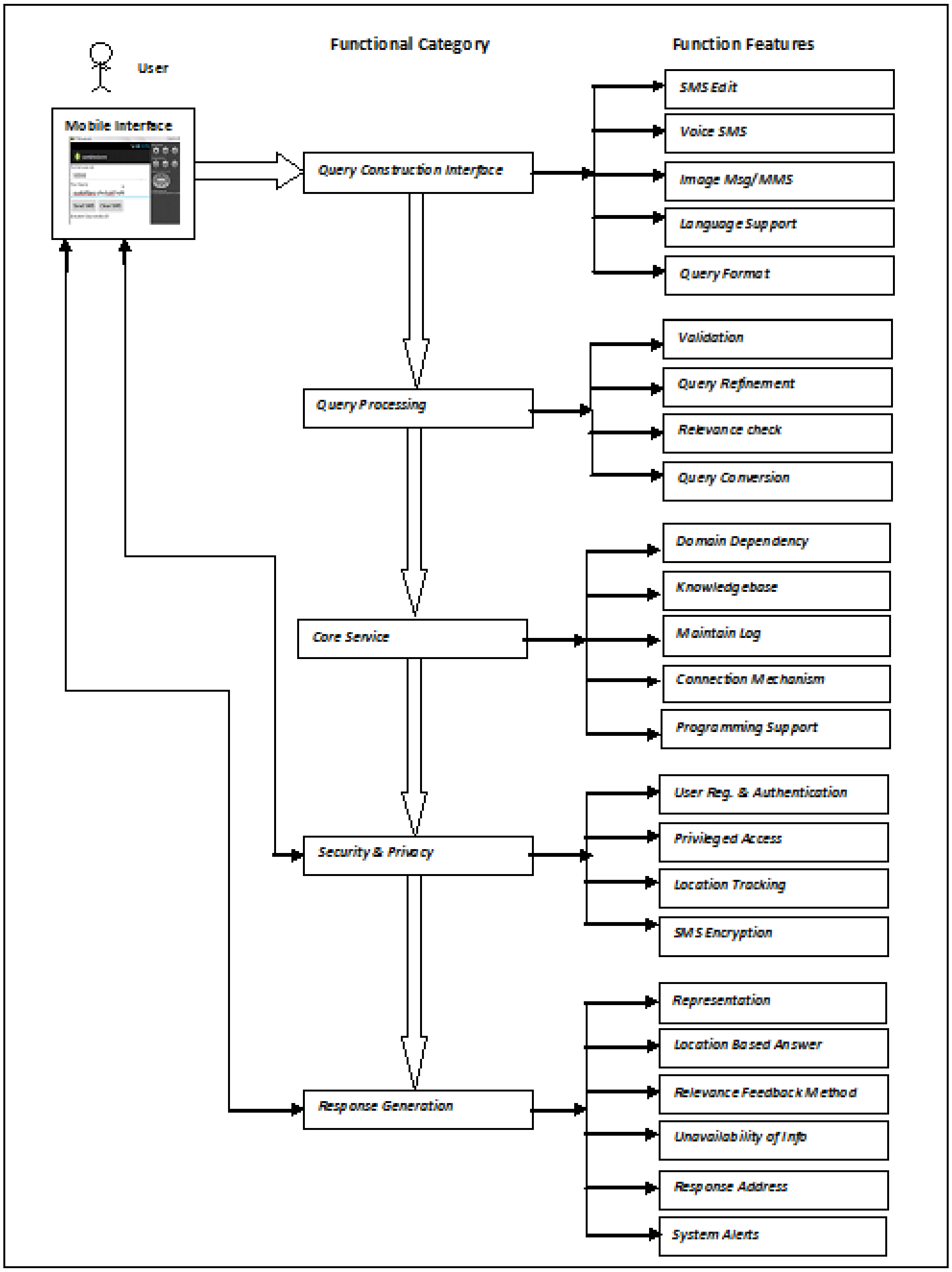

**Figure 1: Functional Model of SMSbIR Systems**

With advancement in scope of IR, Multilingual support has become an important challenge. A set of translators need to use to convert information in one language to another. Google SMS as referred in S14, S27 specifies this feature in restricted form. Cross language information exchange is still part of research. The applications like "Tour Guide" (S28, S26) need to apply multilingual interface to receive wide system acceptance by tourists. A full form of Local language (LL) support on the other hand has concerns such as suitable scripting for writing query and translators to translate the LL statements in standard format. M-governance applications like "Monitoring of polling stations" need to apply LL for effective implementation. Optionally support of scripting and translators could be available with the handset of users or the service software. System S13 for example focuses on this feature.

*D. Query Formats*

Query construction mainly depends on query format defined in the system. A text based query in SMS form needs to include attributes of the information to be retrieved. There are two major methodologies extracted from the reviewed systems.

**Fixed form, planned query-** Systems like S1, S13 apply SMS query in predefined-fixed form. Many of the telemarketing SMS, provide a format that people are asked to use to type the query. This predefined form of query is send as SMS to specific service provider's number. *Service originated SMS query* use this type of format. Such queries are planned and simple but have limited functionality.

**Free form, unplanned query-** *User originated queries* are by and large of free form. User can conveniently write query if there is no any restriction of predefined fixed format. A free form natural language query is thus one of the important demand parameters and a focused objective of most of the research work in this area (S8,S9). System S23 covers a SMSFind project that handles unplanned/flexible NLQ s very effectively and efficiently. It is interesting to study the related work of this system as a special case. Few systems are found with fixed but more flexible query formats. S43 uses soundex phonetics coding NLQ based information access. In S38, keyword patterns are generated to form Natural Language Query with context awareness.

TABLE I. Functional Category : "QUERY CONSTRUCTION INTERFACE"

| Functional Features | Function Points | Category wise distribution of Systems supporting the Functionality for "Query Construction Interface" | | | | | | | | | | | | | | | | | | | | | | | | | | | | | | | | | | | | | | | | | | |
|---|---|---|---|---|---|---|---|---|---|---|---|---|---|---|---|---|---|---|---|---|---|---|---|---|---|---|---|---|---|---|---|---|---|---|---|---|---|---|---|---|---|---|---|---|
| | | Cat A | | | | | | Cat B | | | | | | Cat C | | | | | | | | | | | | | | | | Cat D | | | | | | | | | | | | | | |
| | Systems | 1 | 2 | 3 | 4 | 15 | 32 | 5 | 6 | 10 | 16 | 18 | 24 | 7 | 8 | 9 | 11 | 12 | 13 | 14 | 25 | 26 | 27 | 28 | 30 | 37 | 38 | 41 | 43 | 17 | 19 | 20 | 21 | 22 | 23 | 29 | 31 | 33 | 34 | 35 | 36 | 39 | 40 | 42 |
| SMS Edit | Text editor | √ | | | | | √ | | | | √ | √ | √ | √ | | | √ | √ | √ | √ | √ | | | √ | √ | √ | √ | | | √ | √ | | √ | √ | √ | | | √ | √ | √ | √ | √ | √ | √ |
| | Menu | | | | | | | | | | | | | | | √ | | | | | | | √ | | | | | | | | | | | | | √ | | | | | | | | |
| | Template | | | | | | | √ | | | | | | | | | | | | | | | | | | | | √ | √ | | √ | √ | | | | | | | | | √ | √ | | |
| | Browser | | | | | | √ | | | | | | | | | | | | | | | √ | | √ | | | √ | | | | √ | √ | | | | | | | | | | | √ | |
| Voice Message | IVR | | | | | | | | | | | | | | | | | | | | | | | √ | | | | | √ | | √ | | √ | | | | | | | | | | √ | |
| | Voice SMS | | | | | √ | | √ | √ | | | | √ | | | | | | | | | | | √ | √ | | | | √ | | √ | | | | | | | | | | | | | |
| Image/ MMS | Text MMS | | | | | | | | | | | | | | | | | | | | | | | | | | | | | | | | | | | | | | | | | | | |
| | Image MMS | | | | | √ | | √ | | | | | √ | | | | | | | | | √ | √ | √ | | | | √ | | √ | | | √ | | | | | | | | √ | | | |
| Language Support | Single NL | √ | √ | √ | | | √ | | | | | | | √ | √ | √ | | √ | | | √ | | | | | √ | √ | | | | √ | | | √ | √ | | | √ | | √ | √ | √ | √ | |
| | Multilingual | | | | | | | √ | | | √ | | | | | | | | | √ | | | √ | √ | | | | | √ | | | | | | | | | | | | | | | |
| | Local Language | | | | | | | √ | √ | | √ | | √ | | | | | | √ | | | | | | | | | | | √ | | | | | | | | | | | √ | | | |
| Query Format | Fixed-planned | √ | √ | √ | | | | | | | | √ | √ | | | | | | √ | | √ | | | | | | | | | √ | | | | √ | | | | √ | √ | √ | √ | √ | √ | |
| | Free-form-unplanned | | | | | | | | | | √ | | | √ | √ | √ | | √ | | √ | | | | √ | | √ | √ | | √ | | | | | | √ | | | | | | | | | |

3.2 Analysis of QCI Table

From the Table – I we can extract following important points.

- Most of the systems are working on text messages. Twenty Seven systems have used text based options like text editor, menu and templates for constructing the messages. Remaining systems are using voice/image messages. The related literature of Systems like S2, S3, S4 of Category A, S10 of Category B and S31 of Category D have not specified any reference about the message constructions. The M-Governance systems need to use all the options as mentioned in S5, S6 system's literature.
- Very less work could be recorded on local languages. Only two out of twelve Research Prototype systems and eleven research models could relate this feature. Similar things are found for multiple languages. On the counterpart the M-governance proposed models insist on use of local languages and multiple languages interface. This is necessary for the effective implementations and true benefits of the M-governance applications.
- Maximum systems are using Fixed format planned queries as compared to flexible unplanned queries. Most of the Category D systems (8 out of 15) are using the fixed format option. On the other hand total 11 systems are apply flexible query option out of which 9 systems are of Category C. Natural Language Queries are considered as flexible queries [4] [5].

## 4. QUERY PROCESSING (QP)

The SMS sent by a user is stored in SMS Center (SMSC) of the service provider's server. The text query is transformed into a standard query form. The standard query is used to access relevant information from the knowledge source. Four Functional features pertaining to query processing have been found. Brief information of each functional feature and the details of concerned methodology extracted from the referred systems have been rendered in first subsection. The related cross view of methods vs. systems is depicted in Table – II. This table is evaluated in second subsection.

### 4.1 Functional features of QP

The four functional features of Query Processing category are explained in this subsection. These features are identified as validation, refining, relevance check and query conversion. Let us look at these features and different methodologies involved with each feature.

*A. Validation*

**Validation checks** are used to check if a submitted query is syntactically correct or not. It involves separation of tokens and matching them with a predefined format. The query formats are defined in terms of different key words and their types. Pratima Mitra [9], has used a table including query type, token types and full query format for system S13. This table is maintained to validate every incoming query. Similar method is explained in S1. Relevant **error messages** are sent to the source if some invalid form of message is received. This feature has been extracted from the reference of systems S1, S13, S21, S23, and S25.

*B. Query Refinement*

**Parsing and Normalization** are mainly related to NLQs. They are rather the mandatory steps for refining the query to make it sensible. Parsing includes separation of terms, checking their relevance with the expected query form. Normalization involves eliminating noise or distortion from the received query (S2, S7). New NLP approaches are under progress to resolve the noise in SMS NLQ.

Alexander Ran [4], suggested a method of parsing using semantic tagging. Parsing of the query "*what are the names of my contacts at IBM in Ulm?*" gives <token, class. attribute> pairs as <Ulm, Location.name>, <IBM, Organization.name> etc. It builds a sub-graph to represent abstract meaning of the query.

Two models as "HMM-based spacing model" and "Rule-based Spelling error Correction Model" of query refinement are suggested by Byun, Lee[15]. AiTi Aw, Min hangs, Juan Xiao present "A phrase-based statistical model" for SMS text normalization [18].

The query entered by user could take dispensable form or user may enter some wrong query. Some systems try to repair the query and construct **alternate query.** With semantic checks **noise detection** is done. It is either removed or diluted with normalization. With this a more relevant and meaningful query may be constructed as in systems S9 and S11.

Alternately in case of unanswerable query, alternate query is constructed after dilating some constraints of search (eg. S12). Distorting one or more attributes of the original query is also other way of modifying the search requirements. System S8 applies **tagging** to investigate misspelled or incorrect query terms.

*C. Relevance Check*

Query refinement feature is complemented with relevance check. It supports alternate query formation. Mainly the free form unplanned NLQs need the relevance checks using internal or external ranking. More significant query could be formed by applying ranking. The **ranks/credit-points** are assigned to queries with internal computational mechanism. In case of distorted query send by user, the alternate query is constructed by applying the previous two issues viz. query refinement and alternate query construction. The relevance of alternate query is checked by using **feedback** or **internal ranking** mechanism.

TABLE II. Functional Category : "QUERY PROCESSING"

| Functional Features | Function Points | Category wise distribution of Systems supporting the Functionality for "Query Processing" | | | | | | | | | | | | | | | | | | | | | | | | | | | | | | | | | | | | | | | | | | |
|---|---|---|---|---|---|---|---|---|---|---|---|---|---|---|---|---|---|---|---|---|---|---|---|---|---|---|---|---|---|---|---|---|---|---|---|---|---|---|---|---|---|---|---|---|
| | | Cat A | | | | | | Cat B | | | | | | Cat C | | | | | | | | | | | | | | | | Cat D | | | | | | | | | | | | | | |
| | | 1 | 2 | 3 | 4 | 15 | 32 | 5 | 6 | 10 | 16 | 18 | 24 | 7 | 8 | 9 | 11 | 12 | 13 | 14 | 25 | 26 | 27 | 28 | 30 | 37 | 38 | 41 | 43 | 17 | 19 | 20 | 21 | 22 | 23 | 29 | 31 | 33 | 34 | 35 | 36 | 39 | 40 | 42 |
| Validation | Checks | √ | √ | | | | √ | | | | | | √ | | | √ | | √ | √ | √ | √ | | | √ | | | | | √ | √ | √ | | √ | | √ | | | √ | √ | √ | √ | √ | √ | |
| | Error-Message | √ | √ | √ | | | | | | | | | | | | | | | √ | | √ | | | | | | | √ | √ | √ | | | √ | | | | | | | | | √ | √ | √ |
| Refine-ment | Parsing | | | | | | | | | | | | | √ | √ | √ | √ | √ | √ | √ | | | | √ | | √ | √ | | √ | √ | | | √ | | √ | | | | | | √ | √ | √ | |
| | Normalization | | | | | | | | | | | | | √ | √ | | | √ | | √ | | | | | | √ | | √ | √ | | | | | | √ | | | | | | | | | |
| | Alternate query | | | | | | | | | | | | | √ | | | | √ | | √ | | | | | | | √ | | | | | | | | | | | | | | | | | |
| Relevance check | Ranking | | | | | | | | | | | | | √ | √ | √ | | √ | | | | √ | | | | √ | √ | | √ | | | | | | √ | | | | | | | | | |
| | Feedback Mechanism | | | | | | | | | | | | | √ | | √ | | √ | | √ | | √ | | √ | | √ | √ | | | | | | | | | | | | | | | | | |
| Query Conversion | PDU to Text | | | | | | | | | | | | | | | | √ | | | | | | | | | | √ | | | | √ | | | | | | | √ | | | | | | |
| | PDU to HTML query | | | | | | | | | | | | | | | | | | | | | | | | | √ | | | | | | | | | | √ | √ | | | | | | | |
| | WML | | | | | | | | | | | | | | | | | | | | | | | | | | | | | | | | | | | √ | √ | | | | √ | | √ | |

The feedback assures the user's satisfaction is meeting by the alternate query. The semantic tagging mechanism is suggested by Alexander Ran in his research work for NLQ systems (S8) for matching internal query with available information.

Similar research work has been recorded in the paper [29] of system S23. This system is related to a "SMSfind" algorithm for information extraction in response to the SMS based flexible query send from mobiles. It uses surface pattern matching technique as the extraction method. It applies mean ranking and minimum distance (from hint) as relevance check mechanism.

*D. Query Conversion*

After validation and parsing of the received query in internal format that is acceptable / feasible to the application server is the necessary feature. In most of the related projects it is mentioned that the SMS text is received in Protocol Data Unit (PDU) form. It needs to be converted in to **Plain Text** form (S11, S19, S33) or **HTML query** form (S29, S31, S36) alternatively. This conversion depends upon the compatibility of the application server software and its back end support.

4.2 Analysis of Query Processing.

The Table – II shows all above functional features and the related methodology observed in various systems. The matrix of "Nine Function Points X Forty three systems is evaluated to extract following facts.

- No systems of Category A and Category B seem to care about the query refinement and other counterparts. On the other hands only validation is being applied by each system as a necessary function to check the syntactical formats.

- Category C systems are more working on NLQs and are looking after the issues parsing and noise removal as a major aspect of query processing. A significant scope of research in NLQ SMS could be foreseen from this study.

## 5. CORE SERVICE (CS)

In this section we concentrate on Core service related functional features. Seventeen distinct function points related with Core Service of SMSbISs have been identified and produced in tabulated form as Table-III. As the table shows these seventeen features are categorized in five sub functional categories. All these concerns are discussed in first subsection of this section followed by the analysis of Table-III in next subsection.

### 5.1 Core Service function points

The core service need to hold the complex concerns of the SMSbISs. It is the most complex group of functionality when design and development is concerned. The service providers need to employ a suitable technology backup to build up the whole service mechanism. Then it needs to device special data structures and logic to incorporate the core functionality. All these concerns are characterized by the functional features as domain dependency, knowledge construction, connection mechanism and programming support. Let us look at these features and different methodology involved with each feature.

*A. Domain Dependency*

The core logic of the service is mainly featured by the domain of the application. A system for example may be developed for a specific business. The mobiles could be used as service access terminals by the customers of the business and information access ports by their employees on the other hand. The wireless connectivity may be utilized to overcome the domain dependency by interconnecting services of independent but interrelated services. Crime information of police records may be connected with RTO database to trace the owner of a lost car/vehicle [16].

Practically the core service logic is built upon either single domain or multiple domains. The systems S1 to S3 are restricted to specific application area. The system S14 related to agriculture domain is one example. These are **single domain** systems. The vendor systems such as S4 facilitate for multiple data base connectivity, imparting **multi-domain** IR application. **Domain independent** solutions are part of experimental systems like (S7, S12). In such case the information is delivered, independent of specific domain. A casual information access is the need indeed, promoting a new research trend [8].

*B. Knowledge Base Construction*

Information/data source of the service is built with two options. One is the S**tatic Knowledge Base.** RDBMS like ORACLE, SQL server (S1) hold static data. Another option is **Dynamic Knowledge Base.** The Web based data/information is driven using search engines interface or WAP. The system S8 works on RDF repository based on Ontology. Effective semantic information access using NLQ is supported by such ontology based knowledge construction. Using WEB for SMS based information access has been experimented in system S23. The "SMSfind" algorithm of system S23 has been designed to access search engines to extract relevant documents from web. It extracts relevant snippet from these documents by applying surface pattern matching mechanism [20]. Enterprise resources planner (ERP) software like SAP is another example of Dynamic KB. This alternative is illustrated in systems S2 and S3. Some systems (S12, S14 and S25) may serve with both static and dynamic KBs. The complimentary technical support of GPRS and WAP allow simultaneous access to both types of KBs. **External Knowledge Base**: Some systems may interact with external data base that is of other systems. For example in "Motor vehicle SMS system" (S1), the motor vehicle information is accessed from state crime record, national crime record and database from SP office with a centralized access. It is suggested that RTO database also can be fetched to check the motor vehicle up-to-date information. System S8 also access external data repository in RDF form to answer NLQs. The application of system S3, uses SMS based query to update data base of a dependent service. This is termed as "push information". **Integrated backend:** The architecture of the System S28 experiments the integrated back end in the form of data ware-house. The applications specified in the systems S16, S27, S32 are based on this type of knowledge base. Especially the large scale business processing involves support of integrated backend. Systems like (S14, S26) are coming up with new solutions that apply Value Added Services (VAS) of modern mobile technology (3G). **Multimedia information** accessible on mobiles, allow customers to use mobiles for window shopping, staying comfortably at home.

*C. SMS Log*

Maintaining a log of all incoming and outgoing messages is a mandatory for all SMS based information systems. Some systems apply service provider in*dependent SMS services.* The service gets a special wireless connectivity which is independently identified with a unique service number with lesser number of digits than the normal mobile number. Normally any subscriber's mobile number is around 10 digits, where as the special services are given a 5/6 digit number like '55566'. The SMS send to this service application is first stored in SMS Center (**SMSC**) directly accessible to the system's service. The special systems like mobile based polling/ voting, market feedback, interactive quiz competitions use this type of services.

The service provider dependent system is connected with a standard mobile connection with its PC. In this case the system can organize such SMS log data in its local database. Most of the research models use dependent service and thus need to maintain local SMS log.

It is part of the research now to use such SMS logs for more fruitful mechanism, such as pseudo relevance feedback and data mining. Certain predictions related to customers' perspective, individual or of certain group(s), can be made using this log also can be used to improve knowledgebase and pushing information.

### *D. Connection Mechanism*

To make the SMS based Information system connected to mobiles it needs wireless connectivity. GSM/GPRS network, internet connection and WAP are technologies available to establish PC-Mobile connectivity. SMS gateways / SMS interface can be separately build to execute the mobile service related affairs, keeping the application server free to handle the backend and the service APIs. It is found that the systems as S27 have covered all these options in its related work. The work related to System S17 also depicts the connection using SMSC and SMS interface with receiver's mobile unit. Alternately it specifies using internet server as the other option for the connectivity. The literature related to S39 , S41 systems specify use of these components viz. SMSC, SMS gateway and GSM modems to implement this client and server paradigm.

### *E. Programming Support*

To develop the software as per the system requirement, a suitable programming language is the necessary concern. Though very few papers have specified the language used for software development, this feature has been included as the necessary functional issue for the core systems. JAVA, PHP and Python are the important options extracted from the study. Initially the visual basic language was found to be used in system development based on client server interfacing. Then J2ME based systems are found making the systems platform independent and flexible because of object oriented approach. In latest systems PHP and Python are being used mainly for interfacing mobiles with web based portals offering the information retrieval services. Because of the Android based applications most of the recent developments use JAVA and PHP as the programming languages.

TABLE III. Functional Category: "CORE SERVICE"

| Functional Features | Function Points | Category wise distribution of Systems supporting the Functionality for "Core Service" | | | | | | | | | | | | | | | | | | | | | | | | | | | | | | | | | | | | | | | | | |
|---|---|---|---|---|---|---|---|---|---|---|---|---|---|---|---|---|---|---|---|---|---|---|---|---|---|---|---|---|---|---|---|---|---|---|---|---|---|---|---|---|---|---|---|---|
| | | Cat A | | | | | | Cat B | | | | | | Cat C | | | | | | | | | | | | | | | | Cat D | | | | | | | | | | | | | | |
| | Systems | 1 | 2 | 3 | 4 | 15 | 32 | 5 | 6 | 10 | 16 | 18 | 24 | 7 | 8 | 9 | 11 | 12 | 13 | 14 | 25 | 26 | 27 | 28 | 30 | 37 | 38 | 41 | 43 | 17 | 19 | 20 | 21 | 22 | 23 | 29 | 31 | 33 | 34 | 35 | 36 | 39 | 40 | 42 |
| Domain dependency | Single | √ | | √ | | | √ | | | | | √ | √ | | | | | | | | | | √ | | | √ | √ | | | √ | | | | √ | | √ | | √ | √ | √ | √ | √ | √ | |
| | Independent | | | | | | | | | | | | | √ | | √ | | √ | | | | √ | | | | | | √ | √ | | | | | | | | | | | | | | | |
| | Multiple | | | | √ | | | | | √ | | | | | √ | | √ | | √ | | | | | | | | | | | | √ | | | | √ | | | | | | | | | √ |
| Knowledgebase | Static database | √ | √ | | | | √ | | | | | √ | | √ | | | √ | √ | | √ | | | √ | | | | √ | | | √ | √ | √ | √ | √ | | √ | | √ | √ | √ | √ | √ | √ | √ |
| | Dynamic | | | | | | | | | | | | | | | | √ | | | √ | | | | √ | | √ | √ | | | √ | √ | | | | √ | | | | | | | | | |
| | External | √ | | | √ | | | √ | √ | | | | | | | √ | | | √ | | | √ | | | | | | | | | | | | √ | | | | | | | | | | |
| | Integrated back end | | √ | | | | √ | | √ | | √ | | | | | | | | √ | | | | | √ | | | | √ | √ | | | | | | | | | | | | | | | |
| Maintain Log | SMSC | | | | √ | √ | | | | | | | | | | | | √ | √ | | | | √ | | | √ | | | | √ | √ | | √ | | | | | √ | √ | √ | | √ | | |
| | Local SMS log | | √ | √ | | | | | | | | | | | | | | | | √ | | √ | | | | | √ | | | | | | | | √ | | | √ | | | | | | |
| Connection Mechanism | GSM/GPRS | | √ | √ | √ | | √ | √ | | √ | √ | √ | | | | √ | | | √ | | | | | | | √ | | | | √ | √ | √ | √ | | | | | √ | √ | √ | √ | √ | √ | √ |
| | WAP Portal | | | | | √ | √ | | | √ | | | | √ | | | | √ | | | √ | √ | | | | | | √ | | | | | | | | √ | | √ | | | √ | √ | | |
| | WEB/Internet | | | | | √ | | √ | | √ | √ | √ | | | | √ | | | | √ | | | | | | | | | | √ | √ | √ | √ | | | | | | | | | | | |
| | SMS gateway | √ | | | √ | | | | | | | | | √ | | | | √ | √ | | √ | | √ | | √ | | | | | √ | | | | | | | | | | | | | | |
| Programming Support | JAVA | | | | √ | | | | | | | | | | | √ | √ | | √ | | | | √ | √ | | | | | | | | | | | | | | | | | | | | |
| | PHP | | | | | | | | | | | | | | | | | | | | √ | | | | | | | | | | √ | | | | | | | | | | | | √ | |
| | Python | | | | | | | | | | | | | | | | | | | | | | | | | | | | | | | | | | √ | | | | | | | | | |
| | Other laguages | √ | | | | | | | | | | | | | | | | | | | | | | | | | | √ | | | | | | | | | | | | | | | | |

The functional options in concerned with "knowledge base constructions" are found being ignored in the work of research models. They are mainly handling only static databases. The other systems of remaining categories on the other hands provide good references of the need and applicability of the other options of an effective knowledge base construction.

### 5.2 Observations with respect to CS

When the individual systems are checked for this categorical evaluation the Table – III is constructed. The cross sectional view is the matrix of seventeen function points into Forty three systems. We can extract following facts from this table.

- The functional options in concerned with "knowledge base constructions" are found being ignored in the work of research models. The KB of the systems coming under this category (nine out of eleven) are based on only static databases. The other systems of remaining categories on the other hands provide good references of the capabilities and applicability of the other options such as dynamic KB, integrated and external KB for an effective knowledge base construction.

## 6. SECURITY AND PRIVACY

A secured and protected environment which could be trusted by the customers can only encourage people to practically start using the wireless network for information access. Necessary security and legal measures need to be endorsed to control new form of Cyber Crimes. The truthfulness of information available through SMS must be assured by the service providers or owner of the information. Sundar and Garg [1] have suggested need of security and authentication in m-governance (S5). Kushchu and Kuscu [2] state security and privacy as a prime concern. They say "The government must overcome the mistrust, and assure mobile users that people's privacy is protected". Three major aspects of security and privacy are included in this category. The systems are cross verified for the occurrence of these features in their work to produce the Table – IV. The corresponding issues are discussed in first subsection followed by the analysis with reference of corresponding Table IV in the next subsection. In systems S41, S43 have elaborated the new feature **SMS encryption** for the security purpose of banking and stock exchange data respectively.

### 6.1 Security & Privacy functional features

User registration and authentication, privileged information access and location tracking are the three important concerns we have included in this functional category. Let us first discuss them one by one and then analyze Table IV to understand the research and development in this area.

*A. User Registration and Authentication*

User registration and authentication can be applied as the necessary issue of security. Systems related to academic organizations relate to, "University Examination Systems" (S21, S20, and S30) and paid services like "Telepay" need the security with the support of registration and authentication. Financial systems such as S33, S35 also include this security
At the time of registration users can be allowed to customize their information delivery channel. The service providers can maintain users profile for filtering and customizing information before delivering it to users. This personalization would enhance the privacy measures of the service.

*B. Previleged Access*

If users are asked to register and authenticated, it is possible to restrict access of information as per the privileges assigned at the time of registration based on user's profile. The privacy measures can be effectively employed to enhance the personalized privacy measures resulting in a reliable information channel. System S10 specifies this as its application. Systems S6, S16 reveal need of privacy measures in M-governance applications with specified examples. The systems S16 and S32 have proposed leveled architecture. In such case different level of privileged information access could be managed at different levels. Systems S19, S20, S21 apply **two levels** of privacy privileges like teacher-student. Banking sector (S22), business processes (S2) need **multilevel** privacy privileges to push and pull information.

*C. Location Tracking*

Location based queries in some applications like traffic/ transportation require location tracking. The security measures of SMS based systems may even demand for location tracking.

Tracking of suspects is the concern of police department. System S10 is one of this type of application in m-governance. A real time vehicle tracking based on location tracking is S31. Another system S32 a transportation ticket payment system also needs to apply location tracking mechanism. Another related paper D. Krishna Reddy et. al. specify the need of location identification for security purpose. System S24 an electoral polling activity monitoring system, also needs to look after the security by tracking the sender's field.
There are two major options applied for location tracking. One is Global Positioning systems (GPS) and the other is Cell destination Investigation (CDI).

### 6.2 Security & Privacy table analysis

This Table – IV is built by exploring various methodology (six in total) from the selected Forty three systems. The cross matrix (six into Forty three) presented in this table can be evaluated to gather following facts.

- Very little work is recorded by the researchers under Category C on security and privacy level of functionality. Where the professional and M-governance systems have specified the security and privacy as necessary aspects. Few of these systems (eight out of twelve in total) enforce the registration and authentication as the security issues. Whereas only one out of twelve research prototypes systems have handled its issue. The research models are also being developed with special concern about the security measures. The six out of the eleven systems of this category have been found with these issues as part of their work.
- Location tracking is another area which is totally being overlooked in research models category of SMS based systems (Category D). Similarly only seven systems in total of Forty Three systems (20%) have covered this feature in their work.

## 7. RESPONSE GENERATION (RG).

The processed relevant query in standard form is used to access the information from the server's knowledgebase. The result is then delivered to the designated mobile. The functionality involved in this task, includes five functional features. These features and corresponding methodology (fourteen in total) are discussed in first part of this section followed by the analysis of the Table – V that is constructed by cross checking of the systems against the functionality of this category. It is consisting of a cross matrix of fourteen Methodology into Forty three Systems.

7. 1 RG functional features

Even though the designated task of this functional category seems simple, it has few issues that we have identified in our study. Response representation, Location based answering, relevance feedback mechanism , handling unavailable answer and addressing the recipient(s) are the important features those have been included in this category. These functional features and the related methodology involved with reference of respective systems are now discussed. The tabular analysis of the systems is presented in lateral part of this section.

### *D. Response Representation*

Construction of the response and its delivery to designated end-terminal needs suitable representation. One of the options is the **SMS form** and other possibility is the **e-mail form.** Modern systems S25, S26 send **MMS message** as the system response. Most of the vendor systems facilitate to convert SMS to e-mail and e-mail to SMS (S4). **Multiple messaging** may be required if a query has more than one answer. (S8, S14). In such case message ranking and ordering is the important concern. If answer of a query requires more than 160 characters, it needs to be divided in multiple SMS responses. Sequencing such responses to make them meaningful is not an easy task.

Results may also be generated as **formatted output**. E.g. live information of centre wise voting status of current assembly elections can be broadcasted on TV Screen (S3) in tabular form.

### *E. Location Based Answering*

A location based query can be answered using location tracking mechanism such as GPS. Number of modern age problems is related to location tracking. Numbers of options for technical support of location identification are available.

TABLE IV. FUNCTIONAL CATEGORY : "SECURITY AND PRIVACY"

| Functional Features | Function Points | Category wise distribution of Systems supporting the Functionality for "Security & Privacy" | | | | | | | | | | | | | | | | | | | | | | | | | | | | | | | | | | | | | | | | | | |
|---|---|---|---|---|---|---|---|---|---|---|---|---|---|---|---|---|---|---|---|---|---|---|---|---|---|---|---|---|---|---|---|---|---|---|---|---|---|---|---|---|---|---|---|---|
| | | Cat A | | | | | | Cat B | | | | | | Cat C | | | | | | | | | | | | | | | | | | Cat D | | | | | | | | | | | | | | |
| | Systems | 1 | 2 | 3 | 4 | 15 | 32 | 5 | 6 | 10 | 16 | 18 | 24 | 7 | 8 | 9 | 11 | 12 | 13 | 14 | 25 | 26 | 27 | 28 | 30 | 37 | 38 | 41 | 43 | 17 | 19 | 20 | 21 | 22 | 23 | 29 | 31 | 33 | 34 | 35 | 36 | 39 | 40 | 42 |
| User registration & authentication, | Reg. No. | | | √ | | | | √ | √ | √ | | | √ | | | | √ | | | | | | | | | | | | | √ | √ | √ | √ | | | | | | | | | √ | | |
| | Password | | √ | √ | √ | | √ | √ | √ | √ | √ | | | | | | | | | | | | √ | √ | √ | | | | | | √ | √ | √ | | | √ | | √ | | | | √ | | |
| Privileged access | Two Level | | √ | | | | | √ | | √ | | | | | | | | | | | | | | | √ | | | | | | √ | √ | √ | | | √ | | √ | | | | | | |
| | Multilevel | | | | √ | | √ | √ | | | √ | | √ | | | | | | | | | | √ | √ | | | | | | | | | | | | | √ | | | | | | | |
| Location Tracking | GPS | | | | | | √ | √ | √ | √ | | | | | | | | | | | | √ | √ | √ | | | | | | | | | | | | | | | | | | | | |
| | CDI | | | | | | | | | | | | | | | | | | | | | | √ | | | | | | | | | | | | | | | | | | | | | |
| SMS Encryption | Yes | | | | | | | | | | | | | | | | | | | | | | | | | | | | | | | | | | | | | | | | | √ | √ | |

These Mobile Phone localization techniques can be listed as CID, GPS and GIS. Few systems related to tourism domain (S26, S28) and vehicle tracking (S31) applications need to deliver information related to location tracking. M-governance problems in systems S5, S6, S10 also illustrate good applications of location tracking.

### *F. Relevance Feedback Mechanism*

A reliable IR service is necessary to meet user's satisfaction level. Applying RFM relevant information delivery can be assured. Very few systems are designed, those demand for user's feedback after the response is delivered on user hand set. Demanding to users for ranking the results send by system is the **explicit** method RFM. E.g. Google SMS asks "Did you mean Coffee?", if someone misspells as "cofe" in the corresponding query. This prompts user to spell the word/query sentence correctly and resend it to receive more relevant results (S14). Implicit and blind relevance feedback mechanisms are the other options as mentioned in S23.

### *G. Unavailable Information*

In case of not availability of relevant information in the knowledge base, most of the systems respond messages with a meaning similar to "Not available of data ...Please Contact <Person/URL>" or "Please visit ….for further enquiry". For ex. a System S33 acknowledges user with "Invalid input or No record found" message.

Sunil Kumar Kopparadu et al. [8] solves this problem in S12 by finding out **Next Close Answer** for the unanswered query. e.g. The Query is "Chemical Engineering Thane" .

It cannot be answered as no Engineering college in Thane offers Chemical branch. The system like above would generate another internal alternative query to find Close to answer information. In such case the user can receive one of the responses –

1. $Null for the query, But you can try………..

2. $Computer Engineering – <XYZ Engineering College, Thane W>
3. $Chemical Engineering – <ABC Engineering College, Kurla E>

| Functional Issues | Methodology | Category wise distribution of Systems supporting the Functionality for "Response Generation" | | | | | | | | | | | | | | | | | | | | | | | | | | | | | | | | | | | | | | | | | | |
|---|---|---|---|---|---|---|---|---|---|---|---|---|---|---|---|---|---|---|---|---|---|---|---|---|---|---|---|---|---|---|---|---|---|---|---|---|---|---|---|---|---|---|---|---|
| | | Cat A | | | | | | Cat B | | | | | | Cat C | | | | | | | | | | | | | | | | Cat D | | | | | | | | | | | | | | |
| | | 1 | 2 | 3 | 4 | 15 | 32 | 5 | 6 | 10 | 16 | 18 | 24 | 7 | 8 | 9 | 11 | 12 | 13 | 14 | 25 | 26 | 27 | 28 | 30 | 37 | 38 | 41 | 43 | 17 | 19 | 20 | 21 | 22 | 23 | 29 | 31 | 33 | 34 | 35 | 36 | 39 | 40 | 42 |
| Representation | SMS form | | √ | | √ | | √ | | | | √ | √ | √ | √ | | | √ | | | | | √ | √ | | √ | √ | | | | √ | √ | | | | √ | √ | | | √ | √ | √ | √ | √ | |
| | MMS Form | | | | | | | | | | √ | | √ | | | | | | | | √ | | | | | | | √ | | | | | | | | | | √ | | | | | | |
| | E-Mail | | √ | | √ | | | | | | √ | | | | | | | | | | | | | | | | | | | √ | | | | | | | | | | | | √ | √ | |
| | Multiple Messages | | | | √ | | √ | | | | √ | | | | | | | √ | | √ | | √ | √ | √ | | √ | √ | √ | | √ | | | | | √ | | | | | | | √ | | |
| | Formatted output | | √ | | √ | | | | | | | | | | | | | | | | | | | | | | | | | | | | | | | | | | | | √ | | √ | |
| Location based answering, | Supported | | | | | | | √ | √ | √ | | √ | √ | | | | | | | | | √ | | √ | | | | √ | | | | | | | | | √ | √ | | | | | | |
| Relevance feedback mechanism | Implicit | | | | | | | √ | √ | | | | | | √ | | | | | | | | | | | √ | √ | | | | √ | √ | | | √ | | | | | | | | | |
| | Explicit | | | | | | | √ | √ | | √ | | | √ | | | | √ | | √ | | √ | √ | √ | | | | √ | | | | | | | | | | | | | | | | |
| Unavailability of Information | Null output | | | | | | | | | | | | | √ | | | | | | | | | | | | | | | | | | | | | | | | | | | | | | |
| | Next Close Answer | | | | | | | | | | | | | | √ | | | √ | | √ | | √ | | | | | | | | | | | | | | | | | | | | | | |
| Response Address | Single | √ | | | | | | | | | | √ | | √ | √ | √ | | √ | | √ | | √ | | √ | | √ | √ | | | | | | | √ | √ | | | √ | √ | √ | √ | √ | | √ |
| | Multi-address | | √ | | | | √ | | √ | | | | √ | | | | | | | | √ | | √ | | | | | √ | | | √ | | | | | √ | √ | | | | √ | √ | | √ |
| | Broadcast | | | | √ | | | | | | √ | | | | | | | | | | | | | √ | | | | | | | √ | | | | | | | | | | | √ | | √ |
| | Other application | | | √ | | | √ | | | | √ | | √ | | | | | | | | | | √ | | | | | | | | | | | √ | | | | | | | | | | √ |
| System Alerts | Yes | | | | | | | | | | | | | | | | | | | | | | | | | | | | | | | | | | | | | | | | √ | √ | √ | √ |

*H. Response address*

The relevant information is send as response to destination mobile unit. The response may be meant for single destination or multiple destinations. For example a student's query about his result (S22) may be answered to his/her mobile, but a ragging complaint of a fresher has to generate alert messages to concerned anti-ragging committee members. The present mobile systems support **single address**, **bulk SMS** and even **broadcasting SMS** services. In some applications the responses may be sent as triggers, as information support or as a query to **other application** that offer some dependent service. For example SMS sent as polling information from different polling booths may be processed by the application server to send polling updates to the service of a NEWS channel in the form of real time information support. Live polling counts [18], live cricket scores could be generated by SMS based system servers to be received by the NEWS channels application server for live display. Banking systems also need to deploy this feature for effective interbank transactions.

7.2 RG table analysis

With total 14 Methodology X 43 Systems cross matrix view following facts are gathered to reflect the corresponding open issues.

- Responses in MMS form are the emerging feature that could be extended along with SMS responses to achieve better applicability.
- Only four of twelve the Category-C systems have applied the "next close answer" feature for response generation for an unanswerable query. That means only 11% of the total systems have looked into the problem of unanswered queries while other systems have ignored this problem.
- The relevance feedback mechanism is the important feature that could be included in RG functionality concerns. Categories A systems have not notified this feature in their work. Not much is found about RFM except implicit and/or explicit ranking and credit point assignment. RFM could be explored as the important feature of SMSbIR in all as mobiles could be benefited to have strength of customized and personal information access.

## 7. Functionality Evaluation

The whole functional model has been explored here through the references of different systems of different domains. We have analyzed the tables one by one to comprehend the functionality of the four different system categories. This categorical evaluation is usefull to highlight the gaps between the core and applied research work.

Other perspective of the evaluation is to look at the chronological progress in the research of this field. For this the systems are sorted on base of year of their publications. The chronological order of systems is first analysed with respect to the four categories.

The chronological development in the functionality of SMSbIR systems is analysed with "function point analysis" method [54 ].

### *I. Categorical evaluation.*

If we evaluate the systems in category vs. chronological order as depicted in the Chart-I ,following points could be located

1. Maximmum number of systems are developed at year 2009 those have contributed in development of all categories except the Professional category. Literature of year 2008 shows lot of developments in all four categories.
2. From 2007 onwards no traces of research in m-governance are found. Comparatively more efforts are taken by the developers of Category C and D.
3. In year 2011 no work in Category A and Category B is on record.
4. Not much work is found for the Proffesional systems and M-governance systems after year 2008 and 2009 respectively.
5. Not much progres in this research area is found in year 2002.
6. Only one system has been found of the year 2013. It may be because the actual work is not still available. There fore we exclude this year from further analysis. It only shows the continuety in the research and development in the field through that year.
7. It could be said that remarkable work is being done in the research prototype and research model contributing to the construction of experimental base for the implementation of SMS based Information Systems.

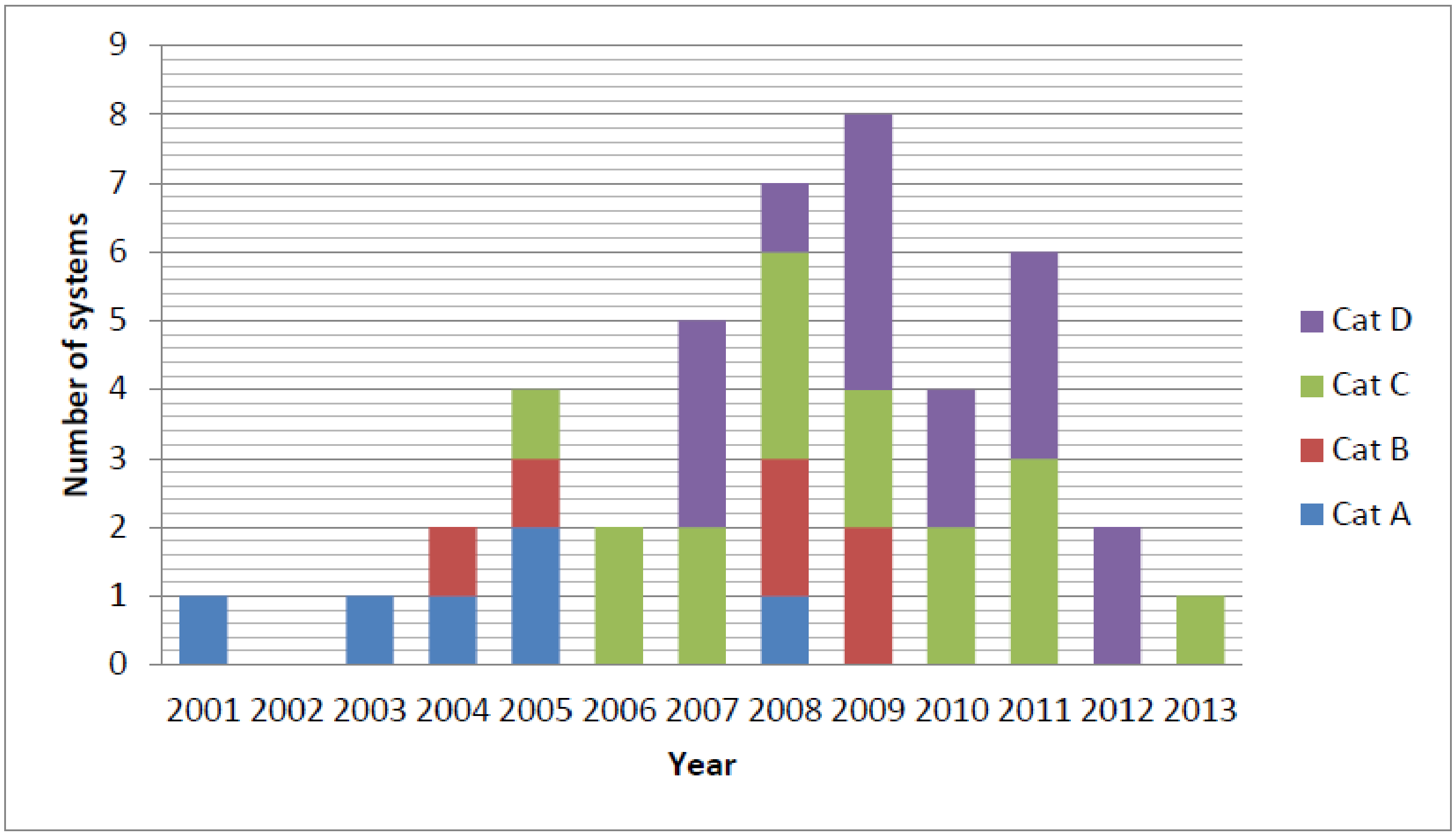

**Figure 2: Chart showing year wise distribution of systems.**

### *J. Function point analysis.*

The functional categories named as query construction interface, query processing, core service, security & privacy and response generation are the five functional categories. The respective tables show that these five functional categories hold the total number of function points as 13, 9, 17, 7 and 15 respectively. To analyze the progress in functionality attended by different system categories is analyzed by arranging systems in categorical manner. The function points of specific functional category attended by the systems are analyzed by computing the sum of average of function point attended by the number of systems of corresponding category.

The function points are quantified by assigning the total number of systems attending the function. This way it is possible to estimate the count of function points, scored by the systems of respective categories Viz. Professional, M-Governance,

Research Prototype and Research Model. This function point analysis is graphically depicted in Figure – 2 (Chart) . The data of function point analysis is attached in the chart's template. Information displayed in this graph can be summarized as follows.

1. Maximum functionality of type Core services is attended by Research Prototype and Research Models.
2. Comparatively, less amount of work is on record with respect to security measures in these categories.
3. Maximum function point counts are scored for the "Core Service" and "Response Generation" functional categories through all the categories.
4. Most of the functional modularity is attended by the Professional category.

## 8. CONCLUSION.

With advancement in mobile technology more and more applications are coming up for effective information access with innovative ideas to utilize its wireless connectivity, improving intelligence in intermediate devices, its popularity irrespective of socio-eco boundaries. These efforts have immerged into a new research trend in the form of "SMS based information retrieval systems". It could be considered as the most fruitful outcome of mobile technology. Users can send a command/ query from their hand sets to access the database, knowledge base, web based data from the service. The wide spread applicability of SMSbIS has introduced significant number of systems throughout the years of last decade. We have selected Forty three such SMSbIS from the published work with the aim of doing comprehensive review to extract all possible function points into a systematic functional model.

Total 63 function points have been extracted from five functional categories including query construction interface, query processing, core service, security & privacy and response generation. These systems are classified into four categories, respectively distinguished as professional systems, m-governance systems, prototype systems and research models. The development in the field is also tracked from year 2001 to 2013. Following open issues can be highlighted from the study of these systems.

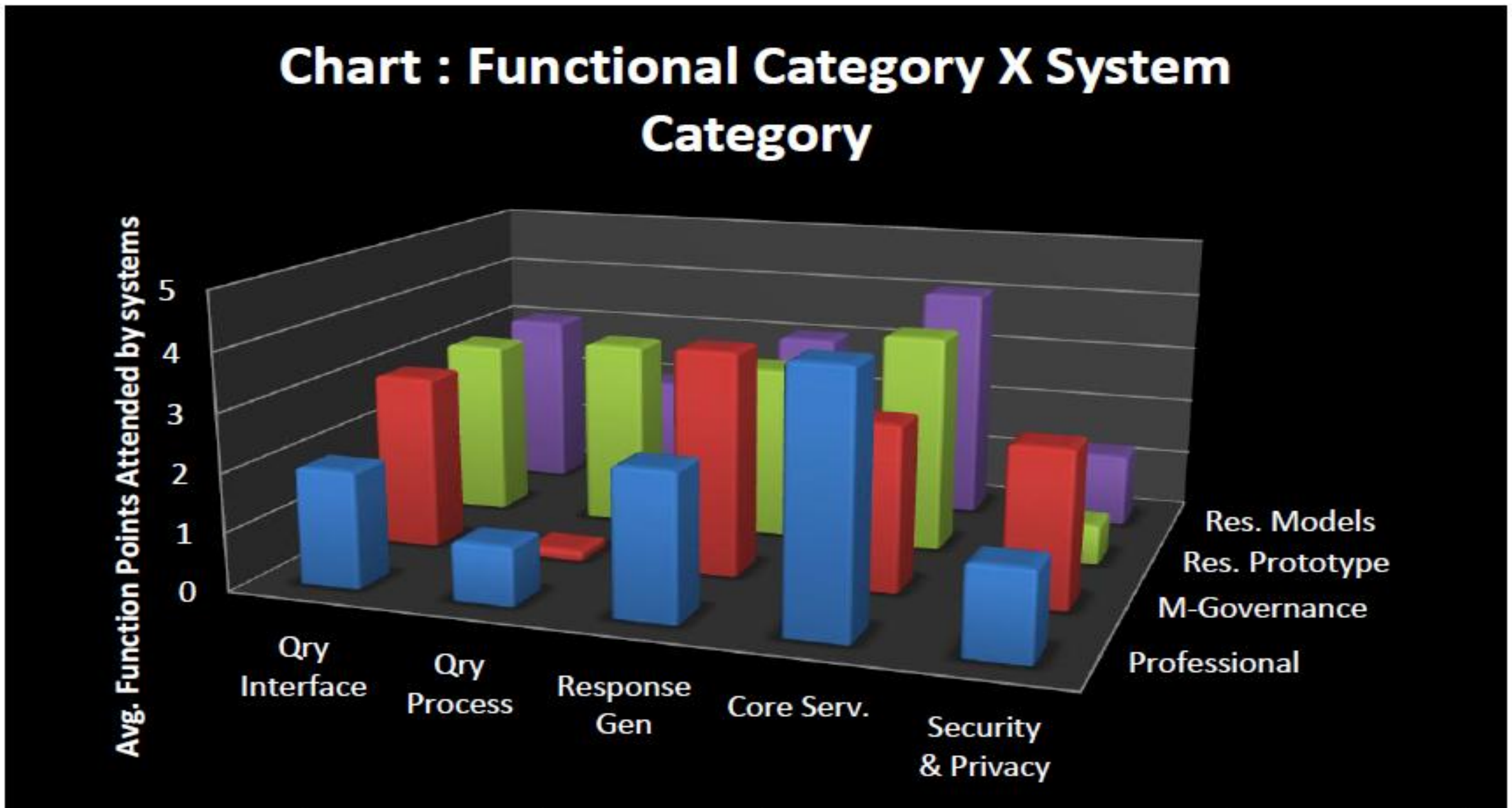

| | Qry Interface | Qry Process | Response Gen | Core Serv. | Security & Privacy |
|---|---|---|---|---|---|
| Professional | 2 | 1 | 2.5 | 4.333333333 | 1.5 |
| M-Governance | 3 | 0.166666667 | 3.833333333 | 2.833333333 | 2.666666667 |
| Res. Prototype | 3.0625 | 3.25 | 3 | 3.75 | 0.6875 |
| Res. Models | 3.066666667 | 2 | 3.066666667 | 4.066666667 | 1.266666667 |

**Figure 3: Chart showing Categorical distribution of SMS based Information Systems with respect to functionality.A. Open Issues**

After examination of the tables of the above functional categories we could find following open issues.

1. The researchers could come up with new ideas for cross language or local language interfacing of SMS queries for information retrieval.

2. A comprehensive work could be done by researchers to frame out various security and privacy measures to make the SMS based IR more widely applicable.

3. The scope of designing personalised information access channels is one of the interesting dimension to this field.

4. The construction of domain independent knowledge base is a challenge that could be assaulted by the researchers.

5. Exploring various options of connectivity with the service and knowledgebase could be explored in detail to find out ideal methodoly for the effective applicability specially in M-governance.

6. Use of SMS log and User profiles for personalised information access on mobiles is an interesting angle of exploration through the strength of this field in order to make SMSbIR as widely accepatable IR solution.
7. Use of flexible queries with flexible natural language and script to access domain independent relevant information on mobiles should be the collective objective of the researchers to make this field of SMSbIR as a reality for the common man by whom the mobiles are being widely used as an all time information exchange medium.

To continue our study in SMS based Information Systems and Information Retrieval we looked for the recent advancements in the research of SMS based information systems and retrieval in recent time span. We could find a few more research papers but because of time and space constraints these recent papers are not included in this review. It seems that till this work is published a few more systems and functional features would be added. Therefore we do not claim about the complete coverage of the functionality of these systems, but this work would definitely help the researchers and developers to resolve their direction of work reducing time of literature review to a great extents.

## REFERENCES-

[1] Sundar, D. K., & Garg, S. (2005, July 10-12). M-governance: A framework for indian urban local bodies. Paper presented at the Proceedings of Euro mGov 2005: The first European Mobile Government Conference, Brighton, UK.

[2] Kushchu, I. and Kushchu, H (2003), "From E-Government to M-Government: Facing the Inevitable", in the proceedings of European Conference on E-Government (ECEG 2003), Trinity College, Dublin.

[3] Govind Kothari, Sumit Negi, Tanveer A. Faruquie " SMS based Interface for FAQ Retrieval" , IBM India Research Lab.

[4] Ran, A., and Lencevicius, R., "Natural Language Query System for RDF Repositories", To appear in *Proceedings of the Seventh International Symposium on Natural Language Processing*, SNLP 2007, 2007.

[5] Rohiza Ahmad, Sameem Abdul-Kareem "A Free-Form Database Query Language for Mobile Phones". Proceedings of the 2009 WRI International conference on Communications and Mobile Computing Volume 03 **,**IEEE Computer Society Washington, DC, USA.

[6] Trimi, S., and Sheng, H. (2008). Emerging Trends in M-Government. *Communications of the ACM*, 51(5): 53-58.

[7] A. Priyadarshani, N. Deepthi, S. Divya, Aswani Kumar "MGuide – A System for Mobile Information System".

[8] Sunil Kumar Kopparadu, Akhilesh Shrvastava, Arun Pande, "SMS based Natural Language Interface to Yellow Pages Directory".

[9] Pramita Mitra, Amitabha Samajpati, Tanmoy Sarkar, Pradip, K. Das "An SMS Based Rural Application for Agricultural Consultancy and Commodity Booking Service"

[10] Schusteritsch, Rudy, Rao, Shailendra and Rodden, Kerry, Mobile Search with Text Messages: Designing the User Experience for Google SMS, In *Proceedings Conference on Human Factors in Computing Systems*, 1777–1780, 2005.

[11] Jeunghyun Byun, Seung-Wook Lee, Young-In Song, Hae-Chang Rim "Two Phase Model for SMS Text Messages Refinement", The AAAI 2008 Workshop on Enhanced Messaging, 2008.

[12] R. Ahmad, S. Abdul-Kareem "Keyword-driven Interface for Mobile Phone's Database Query: A Feasibility Study", in Proceedings of the International Conference on IT and Multimedia, Bangi, Malaysia, 2005.

[13] AiTi Aw, Min hang, Juan Xiao, Jian Su "A Phrase-based Statistical Model for SMS Text Normalization" **.** Association for Computational Linguistics Morristown, NJ, USA.

[14] P. Koehn, F. J. Och , D. Marcu "*Statistical Phrase-Based Translation",* 2003, HLT-NAACL-2003.

[15] Sreangsu Acharyya , University of Texas, Austin , Sumit Negi IBM Indian Research Lab , L.V. Subramaniam IBM Indian Research Lab Shourya Roy IBM Indian Research Lab , "Unsupervised Learning of Multilingual Short Message Service (SMS) Dialect From Noisy Examples".

[16] "Motor Vehicle SMS System", The Centre for Electronic Governance, IIM Ahmedabad **(**www.iimahd.ernet.in*)*

[17] **"**Enterprise SMS Information & Communication System", Bhartia Industries Ltd New Delhi (India), Vendor IG Logix Softech Pvt Ltd, **private published (**www.bchindia.com)

[18] **"**Field Data Collection Over SMS For Live Television News Broadcast" , Vendor IG Logix Softech Pvt Ltd, **private published (**www.imrbint.com).

[19] "spectraSMS Suite , Bringing Purpose to Wireless Technology" **,** Agile Info Systems(AIS) **private published.**

[20] M.M. Soubbotin and S.M. Soubbotin. "Use of patterns for detection of answer strings: A systematic approach". In *Proceedings of TREC*, volume 11. Citeseer, 2002.

[21] Bay Talkitec Combines SMS, 3G Mobile, and Video in New "Video Yellow Pages" Solution Innovative Directory Service Created with SmartCall and Dialogic® HMP Software.

[22] Tony Dwi Susanto, Dr. Robert Goodwin, A/Prof. Paul Calder School of Informatics and Engineering-Flinders University, South Australia, "A Six-Level Model of SMS-based eGovernment" , *published at International Conference on E-Government (ICEG)2008, Melbourne, Australia (Oct 23-24, 2008).*

[23] Tiong T. Goh and Chern Li Liew School of Information Management, Victoria University of Wellington, Wellington, New Zealand "SMS-based library catalogue system: a preliminary investigation of user acceptance".

[24] SMS BASED COACH INFORMATION RETRIEVAL SYSTEM COMMISSIONED ON CENTRAL RAILWAY AND INAUGURATED BY CHAIRMAN, RAILWAY BOARD ON 21ST DECEMBER 2009.

[25] Simon So Hong Kong Institute of Education, Hong Kong, "The Development of a SMS-based Teaching and Learning System".

[26] E. SCORNAVACCA , School of Information Management Victoria University of Wellington, New Zealand, S.L. HUFF & S.MARSHALL University Teaching Development Centre Victoria University of Wellington, New Zealand, "DEVELOPING A SMS-BASED CLASSROOM INTERACTION SYSTEM"

[27] Emmanuel Rotimi Adagunodo Obafemi Awolowo University, Ile-Ife, Osun State, Nigeria , eadigun@oauife.edu.ng Oludele Awodele and Sunday Idowu Babcock University, Ilishan- Remo, Ogun State, Nigeria "SMS User Interface Result Checking System".

[28] Emmanuel Rotimi Adagunodo Obafemi Awolowo University, Ile-Ife, Nigeria , Oludele Awodele Babcock University, Ilishan-Remo, Nigeria , Olutayo Bamidele Ajayi University of Agriculture, Abeokuta, Nigeria. "SMS Banking Services: A 21st Century Innovation in Banking Technology".

[29] Jay Chen New York University jchen@cs.nyu.edu Lakshmi Subramanian New York University lakshmi@cs.nyu.edu Eric Brewer University of California, Berkeley brewer@cs.berkeley.edu "SMS-Based Web Search for Low-end Mobile Devices".

[30] Sanjeev Ranjan, Chief Electoral Officer, Tripura. "SMS based systems for monitoring of Polling Stations: Toward improving Electoral System while considering the range of technical and societal challenges".

[31] MOHD HILMI HASAN, NAZLEENI SAMIHA HARON, NUR SYAFIQAH SYAZWANI MD YAZID Computer and Information Sciences Department Universiti Teknologi PETRONAS Bandar Seri Iskandar, 31750 Tronoh, Perak MALAYSIA. "Development of Multimedia Messaging Service (MMS)-Based Examination Results System".

[32] A.SOUISSI , H.TABOUT and A. SBIHI Department of Telecommunication Systems and Decision Engineering, Ibntofail University, Kenitra, Morocco, "MIR system for mobile information retrieval by image querying".

[33] BEHRANG PARHIZKAR Faculty of Information & Communication Technology, LIMKOKWING University Cyberjaya, Selangor, Malaysia hani.pk@limkokwing.edu.my , ABDULBASIT MOHAMMAD ABDULRAHMAN ALAZIZI Faculty of Information & Communication Technology, LIMKOKWING University Cyberjaya, Selangor, Malaysia Basit5050@yahoo.com. MOHAMMAD NABIL SADEGH ALI Faculty of Information & Communication Technology, LIMKOKWING University Cyberjaya, Selangor, Malaysia Nabil1420@yahoo.com , "PC 2 Phone Event Announcer".

[34] Ma Chang-jie, Fang Jin-yun Laboratory of Spatial Information Technology, Institute of Computing Technology, Chinese Academy of Sciences, Beijing, 100190, PRC - machangjie@ict.ac.cn, "LOCATION-BASED MOBILE TOUR GUIDE SERVICES TOWARDS DIGITAL DUNHUANG".

[35] Mohd Nazri Ismail "Development of WAP Based Students Information System in Campus Environment".

[36] Daciana Illiescu and Evor Hines, School of Engineering, University of Warwick, "SMS Based Student Feedback and Assessment".

[37] D.Krishna Reddy, A.D.Sarma1, Sreenivasa R.Pammi1 and M.V.S.N.Prasad, Chaitanya Bharathi Institute of Technology, Hyderabad R & T Unit for Navigational Electronics, Osmania University, Hyderabad, India. 2Radio and Atmospheric Science Division, National Physical Laboratory, New Delhi , India. E-mail: dkreddi@rediffmail.com , ad_sarma@yahoo.com , "Development of SMS-based G2I System to Generate Reference GIS Maps for Real Time Vehicle Tracking Applications".

[38] Dr. Konstantin Keutner (Siemens AG) Manfred Hildebrandt (Siemens AG) Peter Moritz (Siemens AG) Vinay Yadav (Siemens) AG) A Project co-ordinated by Vincent Blervaque (ERTICO) Tel: +32 2 4000 724 Fax: +32 2 4000 701 Email: v.blervaque@mail.ertico.com ,"Telepayment system for Multimodal Transport Services using Portable Phones".

[39] Mohd Hilmi Hasan and Ahmad Nazrin Ab Manah, Journal of Computing , Volume 3 , Issue 2, Feb 2011 "Enhancing Employee Provident Fund System Through MMS-Based Account Statement".

[40] Fouzia Mousumi and Subrun Jamil, Computer Science and Engineering, University of Chittagong, Bangladesh Chittagong Online Limited, Chittagong, Bangladesh, *International Arab Journal of e-Technology, Vol. 1, No. 3, January 2010, "*Push Pull Services Offering SMS Based m-Banking System in Context of Bangladesh".

[41] Vivian Ogochukwu Nwaocha; National Open University of Nigeria; e-mail:webdevee@yahoo.com, Postgraduate strand at ICTD2010 , "SMS-Based Mobile Learning System: A Veritable Tool for English Language Education in Rural Nigeria ".

[42] M. V. Ramana Murthy, Mobile computing and Wireless Networks CDAC, Electronics city, Banglore "Mobile based Primary Health Care system for Rural India".

[43] Ademola O. Adesina, Kehinde K. Agbele, Nureni A Azeez, Ademola P. Abidoye, "A Query-Based SMS Translation in Information Access System", International Journal of Soft Computing and Engineering ISSN:2231-2307, Volume-1, Issue-5, Nov 2011.

[44] K. Agbele, A. Adesina, N. A. Azeez, A. Abidoye and R. Febba, " A Novel Document Ranking Algorithm that Supports Mobile Healthcare Information Access Effectiveness", Research Journal of Information Technology, ISSN 1815-7432/ DOI: 10.3923/rjjt 2011.

[45] Giorgio Orsi, Letzia Tanca, Eugenio Zimeo, "Keyword-based, Context-aware Selection of Natural Language Query Patterns", EDBT 2011, Uppasala Sweden, ACM 978-1-4503-0528-0/11/0003.

[46] Manoj V. Bramhe, "SMS based Secure Mobile Banking", International Journal of Engineering and Technology, ISSN: 0975-4024, Dec-2011- Jan 2012.

[47] Maria Bibi, Tahira Mahboob, Farooq Arif, "SEIS-SMs based Stock Exchange Information System Using GSM for High Availability and Accessinbility", International Journal of Computer Science Issues, Issue 4, No 3, 2012 ISSN 1694-0814.

[48] Suvarna Nandiyal, Sharada Kadganchi, International Journal of Research in Engineering and Advanced Technology, Vol 1 Issue 4 2013 ISSN : 2320-8791.

[49] Akmal Rakhmadi, Nur Zailan Othman, Ab. Razak Che Husin, "SMS for Complaint Management system", ICITA 2008 ISBN: 978-0-9803267-2-7.

[50] David Pinto, Darnes Vilarino, Yuridiana Aleman, Helena Gomez, Nahun Loya, "The Soundex Phonetic Algorithm Revisited for SMS- based Information Retrieval.

[51] Dr. B. Ramamrthy, S. Bhargavi, Dr. R. Shashikumar, "Development of a Low-Cost GSM SMS-based Humidity Remote Monitoring and Control system dor Industrial Applications.. "International Journal of Advanced Computer Science and Applications, Vol 1 No 4 Oct 2010.

[52] G. Raghavendran , "SMS based Wireless Home Appliance Control System", International Conference on Life Science and Technology 2011 Vol 3.

[53] Victor Olugbemiga MATTHEWS , Emmanuel ADETIBA, "Vehicle Accident Alert and Locator (VAAL)", International Journal of Electrical & Computer Sciences IJECS Vol 11 No. 02, 2011.

[54] A. Abran and P. N. Robillard, "Function Points: a study of their measurement processes and scale tranformations", Journal of Systems and softwares, Vol. 25, No. 2, 1994, pp. 171-184.

## Annexure

## Table VI : Collection of 43 SMSbIR systems

| Category code | Category name | Year of publication | Sys. ID [ref.] | System Description |
|---|---|---|---|---|
| A | Professional Systems | 2004 | S1 [16] | Launched in 2004, IIM Ahmadabad has developed Motor Vehicle SMS system for SP office, Ahmadabad Rural. |
| | | 2001 | S2 [17] | "Bhartia Industries, Ltd", implemented SAP ERP system in 2001 to support its growing business operations including 24X7, information availability to its mobile sales force, channel partners and management team. |
| | | 2005 | S3 [18] | IMRB International solved problem of DD-news channel (24X7) and NIC that enabled them to receive the vote count directly from counting centers into the NIC database server that is directed to DD-news server for direct display on the television screen. |
| | | 2005 | S4 [19] | "spectraSMS", is a mobile messaging product that can work with GSM or SMPP services and can be integrated with RDBMS for B2B or B2C information exchange and processing. |
| | | 2008 | S15 [21] | Bay Talkitec decided and developed a "Video Yellow Pages" application by combining SMS, 3G mobile technology and Video together. The system is de - signed such that, it would respond to a SMS message by pushing a video to the sender's 3G mobile phone over a video call. |
| | | 2003 | S32 [38] | A project report detailing the technical aspects of the common core system architecture of a tele-payment system for multimodal transport services using integrated SMS, MMS, IVR and other Value Added Services (VAS). |
| B | M-governance Systems (MS) | 2005 | S5 [1] | A conceptual framework for mobile governance in urban local bodies in the Indian context with underline set of applications on mobile devices to facilitate the delivery these services. |
| | | 2004 | S6[2] | Presents various aspects for the implementation of m-governance using wireless communication media, with reference of its applicability in current problems. |
| | | 2008 | S10[6] | Reveals the potential in M-government in comparison to e-government with a discussion of the general trends in m-government practices in leading countries. It presents the challenges and issues involved with m-government practices with suitable case studies. |
| | | 2008 | S16[22] | A six level model is presented after investigating the opportunities and popularity of SMS based e-governance as a background of a current research project. The levels are based on the need of dialog between citizen-and government, those are discussed. |
| | | 2009 | S18[24] | Central railway, India has launched its SMS based coach information retrieval system in 2009. The system accepts the coach no from the incoming SMS and sends back the required information of the coach including fault parameters. |
| | | 2009 | S24[30] | Extending mobile tour guide services by means of MMS, SMS, WAP and IVR, it elaborates the location tracking concept of mobiles in detail. |
| C | Prototype Systems | 2009 | S7 [3] | Presents an efficient search algorithm that handles semantic variations in user questions. Users can send their questions using SMS based interface and receive relevant answers on their mobiles. |
| | | 2007 | S8[4] | Developed for natural language question-answer system, it applies effective ranking mechanism for relevance matching of different semantic interpretation with respect to information in RDF repository. |
| | | 2009 | S9 [5] | Presents a new free-form data base query language termed as unplanned query forms, for mobile users. It includes credit point method to check relevance of the answers explicitly. |
| | | 2008 | S11[7] | The system is designed to support SMS based searches for primary services such as hospitals restaurants, tourists places, shopper centers, educational institutes , weather reports and so many. |
| | | 2007 | S12[8] | Proposal of novel interface that enables users to access yellow pages. The central idea of the proposed method is to avoid any constraints on the way the user can query the yellow pages directory information on the mobile by sending a short message services. |
| | | 2006 | S13[9] | The SMS based application for agriculture information system service has been designed for rural people. Users have to send only one SMS to the service center in order to avail any consultancy on crop-specific fertilizers and pesticides or places advance booking for agricultural commodities. |
| | | 2005 | S14[10] | The system is about specialized mobile data services. With the help of the Google SMS team, the authors worked on the iterative design of the service's user experience. The system considers the technical limitations of the SMS standards, user's conceptual model of both SMS and Google Search. |
| | | 2010 | S25[31] | Students send request for their examination results by SMS to the systems, receives the pdf form of the result as MMS. |
| | | 2008 | S26[32] | The system allows users to send MMS query that is a picture of the monument and to receive SMS/MMS as the information of the concerned place. It applies |
| | | | | CBIR as the core concept. |
| | | 2010 | S27 [33] | Suggests use of SMS, phones for the communication amongst the college and staff members. |
| | | 2008 | S28 [34] | Extending tour guide services (MTGS), by means of SMS, MMS, WAP and IVRs. It elaborates the location tracking concept and technology in detail. |
| | | 2006 | S30[36] | The system implemented for Warwick Engineering college, is capable of collecting and processing students' feedback using SMS facility of their mobiles. |
| | | 2011 | S37[43] | This system is again related tio health care. The respective paper reveals the need of SMS translation , normalization of SMS query and ranking of the answerable data with respect to mobile based health care information accesss system. |
| | | 2011 | S38 [45] | The system uses a context based keyword patterns for building SMS based natural language query to access the health related information. It uses ontology for providing the query patterns. |
| | | 2013 | S41 [48] | This system is applying information retrieval by using Image instead of text SMS. The system uses android client to access information about the send image from the server application. Image processing is performed at the server site. |
| | | 2011 | S43 [50] | The system works on Soundex Phonetic Algorithm to device the SMS based information retrieval. It uses the IR database FAQ data set of FIRE 2011 conference for the ecperiments. |
| D | Research Models | 2009 | S17 [23] | Result of investigation of users' cognitive beliefs and intention to use a proposed SMS based library catalogue system is discussed. |
| | | 2009 | S19 [25] | The system is developed to support administrative activities via SMS technology. |
| | | 2007 | S20 [26] | The paper describes the development and trial of a SMS based classroom interaction system. |
| | | 2009 | S21 [27] | Implementation of SMS based result checking system that mainly enforces the security and privacy measures. |
| | | 2007 | S22 [28] | SMS banking system elaborating need of a data mining, knowledge Discovery and Security and Privacy measures in application. |
| | | 2010 | S23 [29] | Design and implementation of "SMSFind", that is an SMS based search system for receive concise single answer for the users' queries across arbitrary topics in one round of interaction. |
| | | 2009 | S29 [35] | Elaborates Wireless Application Protocol (WAP) for the employment of campus information oriented application for mobile terminals. |
| | | 2007 | S31 [37] | Development and applications of a navigation system known as GPS-GSM integrator for real time vehicle tracking. |
| | | 2011 | S33 [39] | The systems has been designed to enhance the PF account system to produce PF account statement in PDF by sending it MMS form over the mobile of the employees. |
| | | 2010 | S34 [40] | SMS based mobile learning is designed to use as an instrumental tool in teaching high school students English language in rural area of Nigeria. The effectiveness of using this new technology for teaching-learning process has been studied by using this system. |
| | | 2011 | S35 [41] | Considering m-Commerce and m-Banking perspective in Bangladesh, a Push Pull services offering SMS (Short Messaging Service) based m-Banking system has been proposed which is able to provide several essential banking services only by sending SMS to bank server from any remote location. |
| | | 2011 | S36[42] | This system is in concern with the health care. The respective paper acknowledges various systems of mobile based health care. The paper discusses primary level of the respective model. The issues like security and privacy are not mentioned. |
| | | 2012 | S39[46] | The system is developed to experiment the security to banking data. It applies AES symmetric cryptographic algorithm to protect the customer transactions from malicious user. |
| | | 2012 | S40[47] | The SMS based stock exchange information system uses GSM to provide centralized database connectivity. The system has defined SMS based instructions to handle the database system using a client server application. |
| | | 2008 | S42 [49] | This system is about handling complaints by applying SMS based management system. This proposed work elaborates how the customer relationaship management can be integrated with the SMS based information system to for more fruitful results. |